\documentclass[10pt,twocolumn,letterpaper]{article}

\usepackage[pagenumbers]{cvpr} %

\usepackage{graphicx}
\usepackage{amsmath}
\usepackage{amssymb}
\usepackage{booktabs}
\usepackage{float}

\usepackage{bm}

\usepackage[pagebackref,breaklinks,colorlinks]{hyperref}

\usepackage[capitalize]{cleveref}
\crefname{section}{Sec.}{Secs.}
\Crefname{section}{Section}{Sections}
\Crefname{table}{Table}{Tables}
\crefname{table}{Tab.}{Tabs.}

\usepackage{xcolor}
\usepackage{enumitem}
\usepackage{xspace}
\usepackage{booktabs}
\usepackage{colortbl}
\usepackage{multirow}
\usepackage{makecell}
\usepackage{lipsum} %
\usepackage{gensymb} %
\usepackage{bm}

\newcommand{\zyqm}[1]{\textcolor{black}{{#1}}}
\newcommand{\gym}[1]{\textcolor{black}{{#1}}}

\newcommand{\Tref}[1]{Table~\ref{#1}}

\newcommand{\fref}[1]{Fig.~\ref{#1}}
\newcommand{\Fref}[1]{Figure~\ref{#1}}

\usepackage[labelsep=period]{caption}
\renewcommand{\paragraph}[1]{\vspace{0.2em}\noindent \textbf{#1}}

\definecolor{MyDarkRed}{rgb}{0.66, 0.16, 0.16}
\definecolor{MyDarkBlue}{rgb}{0.16, 0.16, 0.66}

\newcommand{\MethodName}{GP-NeRF\xspace}
\newcommand{\hashgrid}{hash-grid\xspace}
\newcommand{\Hashgrid}{Hash-grid\xspace}
\newcommand{\densegrid}{dense-grid\xspace}

\def\cvprPaperID{****}	 %
\def\confName{****}
\def\confYear{****}

\begin{document}

\title{Efficient Large-scale Scene Representation with a Hybrid of High-resolution Grid and Plane Features}

\author{
Yuqi Zhang
\qquad
Guanying Chen
\qquad
Shuguang Cui
\\[9pt]
FNii and SSE, The Chinese University of Hong Kong, Shenzhen
}

\twocolumn[{
\renewcommand\twocolumn[1][]{#1}
\maketitle
\vspace{-18pt}
\begin{center}
    \captionsetup{type=figure}
    \includegraphics[width=0.72\textwidth]{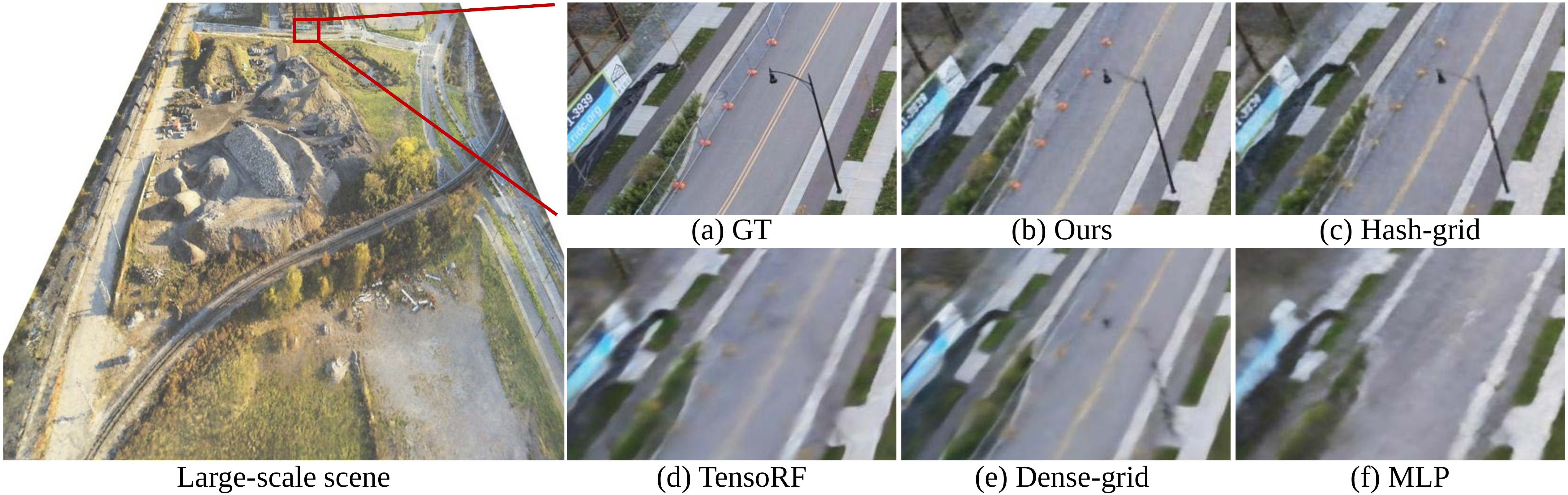}
    \includegraphics[width=0.27\textwidth]{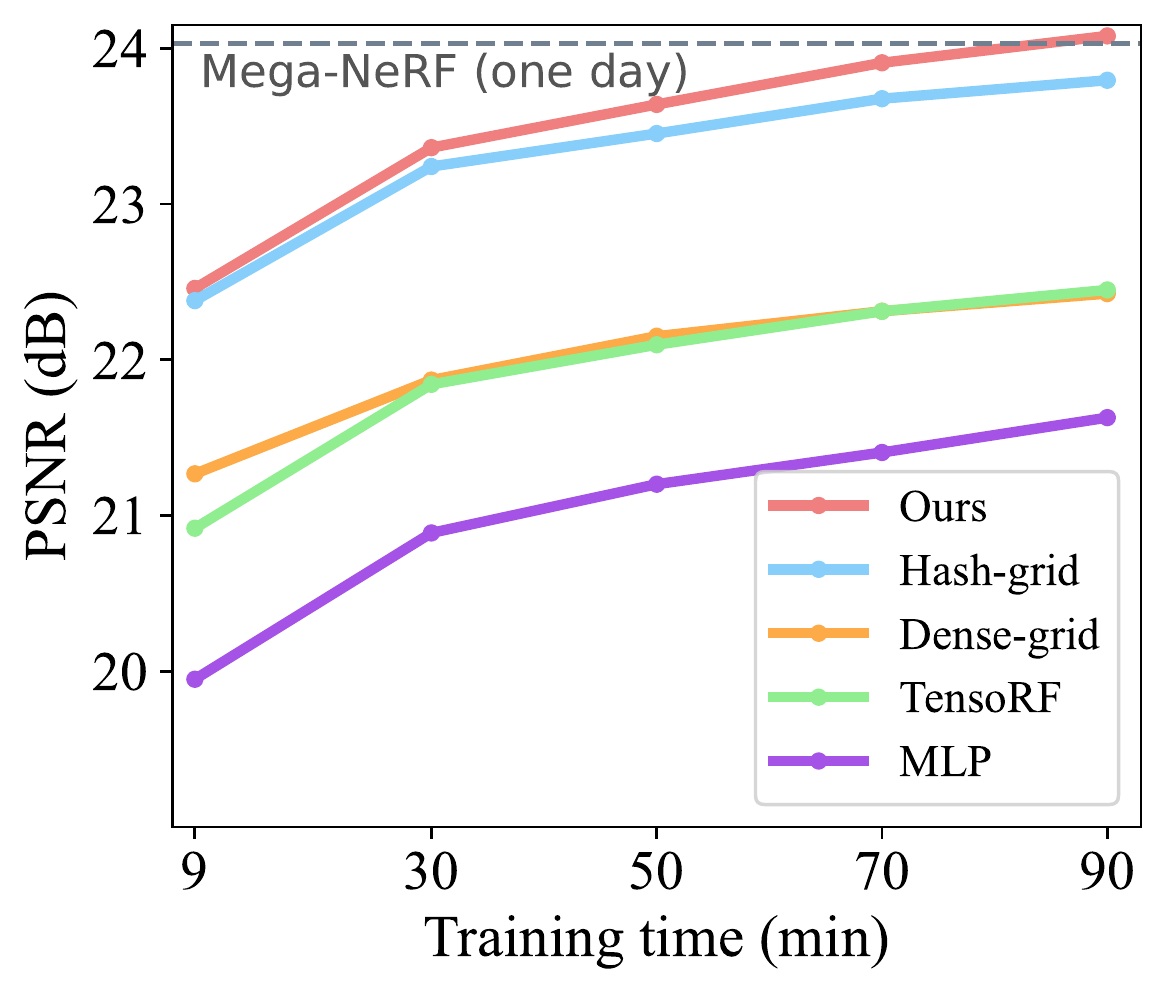}
    \captionof{figure}{We propose a hybrid feature representation for neural radiance fields (NeRF) to enable efficient large-scale \zyqm{unbounded scene modeling}.
    \gym{Compared with results of models that replace the hybrid representation with \hashgrid~\cite{muller2022instant}, \densegrid~\cite{sun2022direct,fridovich2022plenoxels}, TensoRF~\cite{chen2022tensorf} or MLP~\cite{mildenhall2020_nerf_eccv20}, our hybrid representation achieves faster convergence and higher accuracy.}
    Notably, our method can finish training in $1.5$ hours on a single GPU while achieving results better than Mega-NeRF~\cite{turki2022mega} that requires \zyqm{about one day's} training with 8 GPUs.}
    \label{fig:teaser}
\end{center}
}]

\begin{abstract}
    Existing neural radiance fields (NeRF) methods for large-scale scene modeling require days of training using multiple GPUs, hindering their applications in scenarios with limited computing resources. Despite fast optimization NeRF variants have been proposed based on the explicit dense or hash grid features, their effectivenesses are mainly demonstrated in object-scale scene representation. In this paper, we point out that the low feature resolution in explicit representation is the bottleneck for large-scale \zyqm{unbounded} scene representation. 
    To address this problem, we introduce a new and efficient hybrid feature representation for NeRF that fuses the 3D hash-grids and high-resolution 2D dense plane features. 
    Compared with the dense-grid representation, the resolution of a dense 2D plane can be scaled up more efficiently. 
    Based on this hybrid representation, we propose a fast optimization NeRF variant, called \emph{GP-NeRF}, that achieves better rendering results while maintaining a compact model size. 
    Extensive experiments on multiple large-scale \zyqm{unbounded} scene datasets show that our model can converge in 1.5 hours using a single GPU while achieving results comparable to or even better than the existing method that requires \zyqm{about} one day's training with 8 GPUs.
    Our code can be found at \href{https://zyqz97.github.io/GP_NeRF/}{https://zyqz97.github.io/GP\_NeRF/}.
\end{abstract}

\section{Introduction}
\label{sec:intro}

Recent neural scene representation methods have achieved great success in 3D scene reconstruction and novel-view synthesis~\cite{mildenhall2020_nerf_eccv20,yariv2020multiview,niemeyer2020differentiable,sitzmann2020implicit}.
Specifically, neural radiance fields (NeRF)~\cite{mildenhall2020_nerf_eccv20} models the scene with a multi-layer perception (MLP) that takes a 3D point location as well as the view direction as input, and predicts its color and density. The pixel color of a view ray can then be computed by volume rendering using colors and densities of points sampled in that ray.

However, existing methods mainly focus on object-scale scenes. 
Compared with object-scale \zyqm{scenes}, large-scale scenes (\eg, urban-scale environments) can cover more than $1,000,000 m^2$ areas~\cite{turki2022mega}, which often require hours or even days to capture thousands of high-resolution images~(see~\Tref{tab:large_dataset} for examples).

Existing methods for large-scale scene modeling focus on improving the scalability of NeRF. They adopt a distributed training strategy that divides the scene into multiple partitions, each partition is represented by a large MLP model to increase the rendering quality~\cite{turki2022mega,tancik2022block}. 
However, these methods require days of training with multiple GPUs, significantly hindering their applications in users with limited computational resources. 
Recently, many NeRF variants have been proposed to speed up the optimization speed~\cite{sun2022direct,yu2021plenoctrees,muller2022instant}. 
One of the key ideas is to store local features in 3D dense voxel grid~\cite{sun2022direct,yu2021plenoctrees} or \hashgrid~\cite{muller2022instant}, such that most of the time-consuming MLP computation can be replaced with the fast feature interpolations.
However, the effectivenesses of these methods are mainly demonstrated in object-scale scene representation. 

\begin{table}[t]\centering
    \caption{\gym{Statistics of large-scale scene datasets reported in \cite{turki2022mega}.}}
    \label{tab:large_dataset}
    \resizebox{0.48\textwidth}{!}{
    \large
    \begin{tabular}{{l}*{4}{c}}
        \toprule
       Datasets & Resolution & Num of images & Coverage area \\
        \midrule
        Mill19 - Building   & 4608 $\times$ 3456  & 1940 & 262 $\times$ 438 $m^2$\\
        Mill19 - Rubble & 4608 $\times$ 3456 & 1678 & 206 $\times$ 248 $m^2$\\
        Quad 6k         &  1708 $\times$ 1329 & 5147 & 285 $\times$ 420 $m^2$\\
        UrbanScene3D - Residence& 5472 $\times$ 3648 & 2582 &  291 $\times$ 491 $m^2$\\
        UrbanScene3D - Sci-Art  & 4864 $\times$ 3648 & 3019 & 373 $\times$ 317 $m^2$\\
        UrbanScene3D - Campus   & 5472 $\times$ 3648 & 5871 &  1346 $\times$ 1542 $m^2$\\
        \bottomrule
    \end{tabular}
    }
\end{table}

A straightforward idea is to adopt the grid representation to speed up the large-scale scene optimization. 
But as the parameter number of \densegrid representation grows cubically as $O(N^3)$ with resolution $N$, existing methods~\cite{sun2022direct,yu2021plenoctrees} often use a relatively small resolution (\eg, $N=160$ in DVGO~\cite{sun2022direct}) during optimization, making them not suitable for representing large-scale scene\footnote{\eg, if representing a $500\times 250m^2$ scene in a grid with the resolution of $250^3$, each voxel will be responsible for an area of $1\times 2m^2$.}. 
Multi-resolution \hashgrid~\cite{muller2022instant} is an efficient grid structure that applies a hash function to randomly map 3D points into a hash table. By limiting the hash table size (\eg, $T=2^{19}$) in each resolution, the resolution of \hashgrid can be set in a much larger number. However, in the presence of hash collision, directly applying \hashgrid for large-scale scene modeling leads to sub-optimal results.
Therefore, it is important to develop efficient high-resolution feature representation to represent a large-scale scene.

Motivated by the recent success of 2D plane features whose parameter grows quadratically with the resolution as $O(N^2)$ in 3D-aware image synthesis~\cite{chan2022efficient},
in this work, we introduce an efficient \emph{hybrid} representation for large-scale \zyqm{unbounded} scene modeling. 
The key idea of our representation is to enhance the 3D \hashgrid feature with multiple orthogonally placed high-resolution \emph{dense} 2D plane features.  
The structured dense plane features are complementary with the randomly-mapped \hashgrid feature, especially for surface regions with collisions. 
Compared with directly scaling up the hash table size in the \hashgrid representation, \emph{the proposed hybrid representation can achieve higher accuracy with comparable runtime while using much fewer parameters.}
Based on the proposed hybrid representation, we propose an efficient NeRF method, called \MethodName, for large-scale \zyqm{unbounded} scene representation that can finish training with significantly fewer times (see~\fref{fig:teaser}).

To summarize, our key contributions are:  
\begin{itemize}[itemsep=0pt,parsep=0pt,topsep=2bp]
    \item \gym{We propose a new fast optimization NeRF variant}, called \emph{\MethodName}, that is specifically designed for large-scale \zyqm{unbounded} scene modeling.
    \item We introduce a new hybrid feature representation that integrates the complementary features from 3D \hashgrid and dense 2D planes to enable efficient and accurate large-scene modeling.
    \item Experiments show that our method can finish training in $1.5$ hours on a single GPU while achieving results comparable to or even better than the existing method that requires \zyqm{about one day's} training on 8 GPUs.
\end{itemize}

\section{Related Work}
\label{sec:related_works}

\paragraph{Large-scale 3D Reconstruction.}
Large-scale scene reconstruction from multi-view images is a classic problem in computer vision~\cite{agarwal2011building,fruh2004automated,li2008modeling,riegler2021stable,zhu2018very,snavely2006phototourism}. 
Traditional methods often rely on the structure-from-motion (SFM) pipeline to estimate the camera poses~\cite{schonberger2016_sfm_cvpr16}, and apply dense multi-view stereo~\cite{furukawa2010towards,furukawa2010pmvs} to generate the 3D model of the scene. In this work, our method adopts the recent neural representation~\cite{mildenhall2020_nerf_eccv20} for scene reconstruction and novel-view synthesis.

\paragraph{Neural Scene Representation.}
Neural scene representation has revolutionized the problem of scene reconstruction and novel-view synthesis~\cite{shum2000review,tewari2020state,tewari2021advances,xie2022neural}.
In particular, neural radiance fields (NeRF)~\cite{mildenhall2020_nerf_eccv20} has attracted considerable attention for its photo-realistic render quality. Many follow-up methods are then developed to improve its robustness~\cite{Zhang20arxiv_nerf++,martin2021_nerfw_cvpr21,barron2021mip}, generalization~\cite{wang2021_ibrnet_cvpr21,chen2021mvsnerf,yu2021_pixelnerf_cvpr21,xu2022point,trevithick2021grf}, dynamic modeling~\cite{pumarola2021_dnerf_cvpr21,tretschk2020_nonrigid_arxiv,li2021_nsff_cvpr21}, scene decomposition~\cite{zhang2021nerfactor,boss2021nerd,yang2022ps}, etc.

\paragraph{Large-scale Scene Rendering.}
There are methods extending NeRF for handling large-scale scenes. NeRFusion~\cite{zhang2022nerfusion} proposes a progressive update scheme for large indoor scene reconstruction by gradually constructing the local volumes to build the final global volumes. 
Urban Radiance Fields~\cite{rematas2022urban} makes use of the LIDAR data and RGB images to achieve better reconstruction results for street-view environments. 
BungeeNeRF~\cite{xiangli2022bungeenerf} employs the residual learning strategy to train a NeRF-based model for rendering multi-scale data of a city from Google Earth Studio.

Some methods adopt a divide-and-conquer strategy to handle large-scale scenes.
Block-NeRF~\cite{tancik2022block} decomposes the street views of a city into several blocks, and each block is represented by an individual NeRF.
Mega-NeRF~\cite{turki2022mega} divides the space according to the camera distributions to render a large-scale scene captured by a drone.
Similarly, Wu~\etal~\cite{wu2022scalable} stack multiple tiles (each tile consists of two MLPs) based on a global mesh proxy for scalable indoor scene rendering.
However, these methods merely consider the ability to reconstruct large scenes, and suffer from a long training time and low efficiency.
Note that the divide-and-conquer strategy can also be integrated with our method.

\paragraph{Efficient NeRF Rendering and Optimization.} 
To improve the rendering speed of NeRF, some follow-up works have been proposed~\cite{liu2020neural,yu2021plenoctrees,wu2022diver,wizadwongsa2021nex,neff2021donerf,hedman2021baking,lombardi2021mixture,lin2022efficient,wang2022fourier}.
NSVF~\cite{liu2020neural} constructs a sparse voxel field, in which each vertex of the voxel stores local attributes. 
PlenOctrees~\cite{yu2021plenoctrees} utilizes spherical harmonics instead of directly predicting RGB values, and adopts an Octree-based data structure to accelerate the rendering speed. 
KiloNeRF~\cite{reiser2021kilonerf} decomposes the NeRF into thousands of MLPs.
Besides, FastNeRF~\cite{garbin2021fastnerf} factorizes the original NeRF and leverages caching to speed up rendering.

Recently, several approaches for fast optimization of NeRF have been proposed~\cite{sun2022direct,fridovich2022plenoxels,muller2022instant,chen2022tensorf,gao2022reconstructing,guo2022ndvg}. 
DVGO~\cite{sun2022direct} and Plenoxels~\cite{fridovich2022plenoxels} adopt the explicit voxel-grid representation to replace the time-consuming MLP queries with the fast grid interpolation operation. 
Instant-NGP~\cite{muller2022instant} further encodes the feature of a 3D point by multi-resolution hash encoding to support high-resolution grids with low memory consumption. 
In addition to grid-based methods, there are some other works on fast optimization.
EfficientNeRF~\cite{hu2022efficientnerf} accelerates the training by removing redundant sample points based on the coarse reconstruction.
TensoRF~\cite{chen2022tensorf} represents the radiance fields as 4D tensor fields and utilizes the tensor factorization to speed up the training phase.
EG3D~\cite{chan2022efficient} proposes a tri-plane representation to enable high-resolution multi-view consistent image generation.
\gym{Concurrent to our work, some method utilizes plane features for dynamic scene modeling~\cite{shao2022tensor4d,fridovich2023k,cao2023hexplane}. In contrast to these methods, we focus on enhancing the \hashgrid feature with high-resolution plane features to enable efficient large-scale unbounded scene modeling.}

\section{Preliminary}%

\newcommand{\point}{\boldsymbol{x}}
\newcommand{\pointcolor}{\boldsymbol{c}}
\newcommand{\density}{\sigma}
\newcommand{\viewdir}{\boldsymbol{d}}
\newcommand{\nerfmlp}{\mathrm{MLP}}
\newcommand{\nerfsmall}{\nerfmlp_{\mathrm{small}}}
\newcommand{\raycolor}{C}
\newcommand{\ray}{\boldsymbol{r}}
\newcommand{\explicitgrid}{\mathcal{G}}
\newcommand{\feature}{\boldsymbol{f}}
\newcommand{\resolution}{N}
\newcommand{\featnum}{F}

\paragraph{Neural Radiance Fields.} 
Given a set of images with calibrated camera poses, NeRF~\cite{mildenhall2020_nerf_eccv20} represents the scene through the weights of a multilayer perceptron (MLP). 
Taking a 3D point position $\point_i = (x, y, z)$ and the view direction $\viewdir$ as input, the NeRF MLP can predict the corresponding density $\density_i$ and color $\pointcolor_i$ of the input:
\begin{align}
    \label{eq:nerfmlp}
    (\density_i, \pointcolor_i) & = \nerfmlp (\point_i, \viewdir).
\end{align}
To render the pixel color of a view ray, multiple points are sampled along the ray and fed into the MLP to query their densities and colors. Volume rendering~\cite{kajiya1984ray} is then apply for accumulating the discrete color values to compute the pixel color:
\begin{align}
    \label{eq:volume rendering}
    \hat{\raycolor}(\ray) = \sum_{i=0}^{N-1} T_i (1 - \exp( -\density_{i} \delta_{i})\pointcolor_i),
\end{align}
where $T_i = \exp( -\sum_{j=0}^{i-1} \density_j \delta_j)$ denotes the accumulated transmittance and $\delta_i$ denotes the distance between adjacent sample points. 

\paragraph{Fast Feature Interpolation for Acceleration.}
To produce high-quality rendering, NeRF adopts a large MLP model to represent the scene (\ie, $9$ fully-connected layers with a channel number of $256$). As millions of MLP queries are needed in each iteration, a long optimization time is required, especially for large-scale scenes.

To speed up the training, methods based on the explicit representation are proposed~\cite{sun2022direct,fridovich2022plenoxels,chen2022tensorf,muller2022instant}. These methods replaced most of the time-consuming MLP computation with a fast feature interpolation operation on the explicit representation $\explicitgrid$ (\eg, a dense voxel grid or \hashgrid). Taking the interpolated feature $\feature$ as input, a small MLP model can be applied to regress the density and color, significantly reducing the optimization:
\begin{align}
    \label{eq:nerfmlp}
    \feature_i & = \mathrm{Interpolate}\ (\explicitgrid, \point_i), \\
    (\density_i, \pointcolor_i) & = \nerfsmall \ (\feature_i, \viewdir).
\end{align}

As the parameter number of the voxel grid grows dramatically with the increase of resolution $\resolution$, these methods adopt a small resolution (\eg, $160$ in DVGO~\cite{sun2022direct} and $300$ in TensoRF~\cite{chen2022tensorf}). 
Although working well on object-scale scenes, the problem of low-resolution grid features become noticeable in a large-scale scene.
Although the \hashgrid representation~\cite{muller2022instant} can set a much larger resolution by fixing the hash table size, the collision effect will become more severe.
Therefore, it is important to develop efficient high-resolution feature representation for fast optimization of large-scale scenes.

\begin{figure*}[th] \centering
    \includegraphics[width=0.9\textwidth]{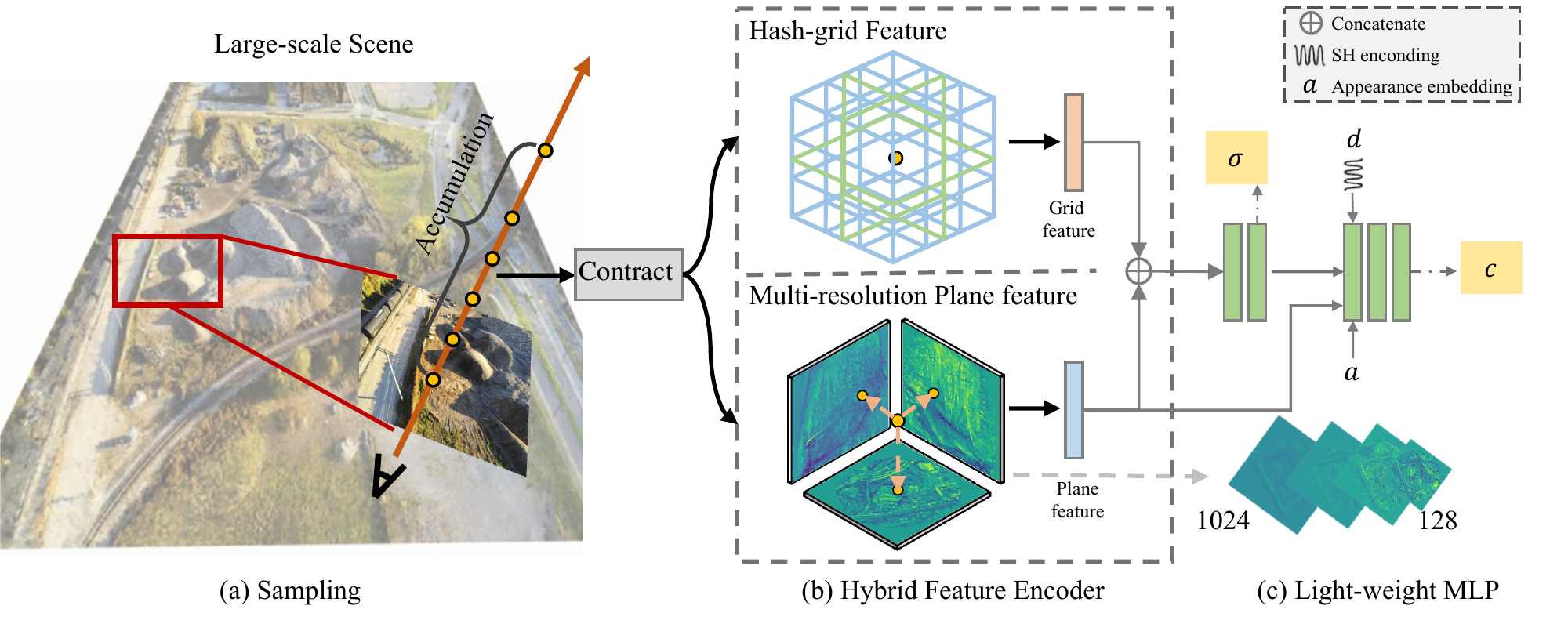}
    \caption{\textbf{Overview.} First, we sample 3D points along the rays emitted from pixels. Second, we parameterize the space into a contracted space for a compact space parameterization. 
    Then, querying the position into the proposed hybrid-representation to extract \hashgrid feature and multi-resolution plane features, which are fed into
    a light-weight MLP to predict the sigma $\sigma$ and the color values $c$ (color MLP requires the additional view directions, appearance embeddings, \zyqm{and plane feature input}). 
    Finally, the image colors can be computed by volume rendering.} \label{fig:method_overview}
\end{figure*}

\section{Method}%
\label{sec:method}
Our goal is to develop an efficient high-resolution feature representation to improve the capacity of fast optimization NeRF methods in representing a \emph{large-scale} scene.

We start by revisiting the recent efficient hash-grid~\cite{muller2022instant} representation for scene modeling and identify its strengths and limitations.
Based on the discussion, we introduce an efficient \emph{hybrid} representation that enhances the 3D \hashgrid feature with high-resolution \emph{dense} 2D plane features, just with a small increase in the parameters and runtime.
Then, we parameterize a large-scale unbounded scene in a contract space~\cite{barron2022mip} to enable a compact space representation.
Last, we introduce the network architecture and optimization strategy~(see~\fref{fig:method_overview}).

\subsection{Multi-resolution Hash-grid Representation} 
Multi-resolution hash-grid~\cite{muller2022instant} is an efficient data structure to represent a scene with high-resolution grids (\eg, $1024$ or even higher resolution) without significantly increasing the parameter number by randomly mapping 3D points to a linear hash table with a fixed size.
The parameter number of a multi-resolution \hashgrid is bounded by $L\cdot T\cdot \featnum$, where $L$ is the number of the resolution, $T$ and $\featnum$ are the hash table size and feature dimension in each resolution. The suggested configuration is ($L=16, T=2^{19}, \featnum=2$) to balance the trade-off between the capacity and efficiency~\cite{muller2022instant}, leading to a bound of $2^{24}$ parameters.

Our experiment shows that the \hashgrid representation with a highest resolution of $2048$ clearly outperforms the low-resolution \densegrid~\cite{sun2022direct} and TensoRF~\cite{chen2022tensorf} representation in large-scale scene rendering, demonstrating the resolution of feature resolution plays a very important part in large-scale scene modeling. 
However, due to the collision problem in random hash mapping, the interpolated feature inevitably contains mixed information for different surface points, limiting the performance of NeRF model.
We also observe that increasing the hash table size $T$ can improve the results, but at the cost of a significant increase in parameter number and a longer optimization time~\cite{muller2022instant}.

This discussion motivates us to introduce an efficient strategy to boost the \hashgrid feature for large-scale scenes just using a small number of parameters and runtime.

\subsection{High-resolution 2D Plane Features}%

Observing that 2D plane features whose parameter grows quadratically with the resolution as $O(N^2)$ can provide strong information to enable 3D-aware image-synthesis~\cite{chan2022efficient} and 3D reconstruction~\cite{peng2020convolutional},  we propose to enhance the \hashgrid feature with high-resolution dense plane features for large-scale scene representation.

\paragraph{Orthogonally Placed 2D Plane Features.}  
We design the plane features as three orthogonally placed planes with a resolution of $N$ and a feature dimension of $\featnum$, then the parameter number for each plane is $N^2\cdot \featnum$.
For a queried 3D point, we first orthogonally project it on these three planes, and obtain the 2D plane features with bilinear interpolation. 
We then concatenate the three interpolated features to form a feature vector of length $3\cdot \featnum$.

We are able to scale up the resolution of the dense 2D planes to $1024$. 
By doing this, our plane representation can provide features in a resolution that ``equivalent to'' a \densegrid with $2^{30}$ parameters, which is difficult to achieve in a \densegrid due to the significant memory consumption. 

\paragraph{Efficient Multi-resolution Design.} 
However, the feature dimension cannot be too large as the parameters will increase sharply (\eg, setting $\featnum=8$ already results in a parameter number of $8\times1024\times1024=2^{23}$), or too small (\eg, $\featnum=2$) as the plane features will not provide useful information (see our ablation study \zyqm{in \Tref{tab:feature dim and multi-scale}}).
To fulfill the goal of providing high-resolution dense features, we design multi-resolution planes following~\cite{muller2022instant}.
Specifically, we adopt a four resolutions configuration $N\in \{128, 256, 512, 1024\}$, each has a feature dimension of $2$, resulting in an $8$ dimension multi-resolution features.
This design can effectively boost the \hashgrid feature while maintaining a low parameter numbers.

Moreover, as in a large-scale scene, the height of a scene is often smaller than the horizontal length. To reduce the waste of features in the vertical planes, we scale the vertical planes using the camera altitude measurements.

\paragraph{Relation to Previous Work.}
Our method is inspired by EG3D~\cite{chan2022efficient} which proposes a tri-plane representation, where each plane has a shape of $256\times 256\times 32$, and the interpolated features from three different planes are fused by an add operation for GAN-based image generation.
In contrast, our goal is to design efficient high-resolution features for large-scale scene modeling by scaling up the plane feature to $1024$ and using a multi-resolution strategy to maintain a low parameter numbers. 
Moreover, we use a simple concatenation operation to combine the three interpolated features, as we experimentally find that applying the add operation leads to worse results, which might be explained by the small feature dimension in our representation (\ie, $8$ in ours and $32$ in EG3D). 

The recent TensoRF~\cite{chen2022tensorf} decomposes the radiance fields into several matrices and vectors. 
However, the matrix has a shape of $N^2\cdot R\cdot F$, where $R=192$ is the component number. Compared with the shape of $N^2\cdot F$ in our plane feature, the matrices in TensoRF representation cannot be efficiently scaled to a high resolution (\eg, $1024$), making it not suitable for large-scale scene modeling.

\subsection{Hybrid of Grid and Plane Features}
Given a queried 3D point, we interpolate the \hashgrid to get a $32$-dimension grid feature and the orthogonally placed planes to get a $24$-dimension feature.
Then these two features are concatenated to form a hybrid feature to be the input of NeRF MLP.
Note that there are more sophisticated feature fusion strategies (\eg, MLP fusion) to fuse the \hashgrid and plane features.
However, to maintain a simple and efficient architecture and also to better demonstrate the effectiveness of our plane features, we just use a simple concatenation.

Compared with the \hashgrid that adopts a random mapping hash-function, the orthogonally placed plane representation introduces a structured and high-resolution dense feature, which is complementary to the \hashgrid feature, especially in regions with hash collision.  As a result, the proposed hybrid feature representation can provide better features for later density and color prediction.

\subsection{Scene Space Parameterization}%
\label{sub:space parameterization}

As a large-scale scene covers a wide spatial range and is always unbounded, it is important to parameterize it in a way that it can be represented by a grid-like data structure.

We adopt the mip-NeRF 360 parametrization~\cite{barron2022mip,sun2022improved} to represent a unbounded scene.
Considering a scene whose center is the origin, 
we divide the scene into foreground and background regions, separating by a pre-define bound $B$. 
Given a 3D point, we will first normalize it by $\bm{x} = \bm{x} / B$, and then applied the space contraction:
\begin{align}
            \label{eq:contract space}
            \point = \begin{cases}
    \point , & \|\point \|_p \leq 1~; \\
    \left(1+b-\frac{b}{\|\point\|_p}\right) \left(\frac{\point}{\|\point\|_p}\right) , & \|\point\|_p > 1 ~, \\ 
\end{cases}
\end{align}
where $p$ denotes $p$-norm, $b$ is to control the size of background space. Following~\cite{barron2022mip}, we set $p=2$ and $b=1$. 

\begin{figure}[t] \centering
    \includegraphics[width=0.48\textwidth]{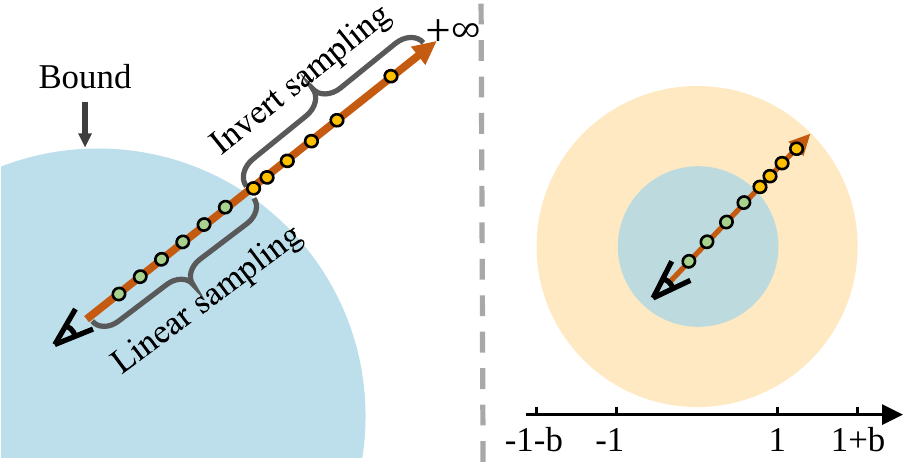}
    \\
    \vspace{-0.3em}
    \makebox[0.235\textwidth]{\footnotesize (a) Euclidean space}
    \makebox[0.235\textwidth]{\footnotesize (b) Contracted space}
    \\
    \caption{Illustration of the scene parameterization. According to the pre-defined bound, we adopt linear sampling and invert linear sampling for the foreground and background areas, respectively. Then, the sampled points are contracted into a compact space.} \label{fig:contraction}
\end{figure}

By doing this, the distribution of points is more compact, where the foreground point (\ie, $\|\point \|_p \leq 1$) is unchanged and a point at infinity will be mapped to the sphere with a radius of $1+b$.
This compact representation enable us to represent the scene with the proposed hybrid feature representation (see~\fref{fig:contraction}).

\subsection{Model Optimization} 
After defining the hybrid representation and scene parameterization for large-scale scenes, we introduce our network architecture and optimization strategy.

\paragraph{Network Architecture.}
Given a 3D point position $\point_i$ and its direction $\viewdir$ as input, our method first extract the features from the hybrid feature representation. 
These hybrid features are then fed into a 64-channel two-layer MLP network that regresses the view-independent density value $\sigma_i$ and a 15-channel geometry feature.

Then, the view direction encoded with the spherical harmonics function, the per-frame appearance embedding~\cite{martin2021_nerfw_cvpr21}, \zyqm{and the plane feature are concatenated with the geometry features} to be the input of a 64-channel three-layer network to predict the view-dependent color value. 
The per-frame appearance embedding is adopted to account for the illumination changes in the captured images. \zyqm{We experimentally found that include the plane feature to the color MLP leads to better results, as discussed in the supplementary material.}

\paragraph{Point Sampling.}
We adopt a linear sampling strategy for foreground and an invert sampling strategy for the background~\cite{sun2022improved} (see~\fref{fig:contraction}), which satisfies the need of high-quality foreground rendering and covers a wide range of background. 
In details, we sample $N_{f}=128$ and $N_{b}=64$ points per-ray for these two areas, respectively. 
We also adopt the coarse-to-fine sampling strategy~\cite{mildenhall2020_nerf_eccv20}, where the coarse and fine stages have the same number of sampled points, but use the same model to reduce the model size.

\label{sec:Experiments}
\begin{table*}[t] \centering
    \newcommand{\Frst}[1]{\textcolor{red}{\textbf{#1}}}
    \newcommand{\Scnd}[1]{\textcolor{blue}{\textbf{#1}}}
    \caption{\gym{{\bf Quantitative comparison} on the testing datasets. * indicates that Mega-NeRF's results are computed with its released models}.
    }
    \label{tab:quantitative}

\makebox[\textwidth]{} 
\resizebox{\textwidth}{!}{
\begin{tabular}{l||*{3}{c}|*{3}{c}|*{3}{c}|*{3}{c}|*{3}{c}|*{3}{c}||c}
    \toprule
    & \multicolumn{3}{c|}{Mill 19 - Building} 
    & \multicolumn{3}{c|}{Mill 19 - Rubble}
    & \multicolumn{3}{c|}{UrbanScene3D - Residence}
    & \multicolumn{3}{c|}{UrbanScene3D - Sci-Art}
    & \multicolumn{3}{c|}{UrbanScene3D - Campus}
    & \multicolumn{3}{c||}{Quad 6k}
    & \multicolumn{1}{c}{Average}\\
    Method
    & \multicolumn{1}{c}{$\uparrow$PSNR} 
    & \multicolumn{1}{c}{$\uparrow$SSIM} 
    & \multicolumn{1}{c|}{$\downarrow$LPIPS} 
    & \multicolumn{1}{c}{$\uparrow$PSNR} 
    & \multicolumn{1}{c}{$\uparrow$SSIM} 
    & \multicolumn{1}{c|}{$\downarrow$LPIPS} 
    & \multicolumn{1}{c}{$\uparrow$PSNR} 
    & \multicolumn{1}{c}{$\uparrow$SSIM} 
    & \multicolumn{1}{c|}{$\downarrow$LPIPS} 
    & \multicolumn{1}{c}{$\uparrow$PSNR} 
    & \multicolumn{1}{c}{$\uparrow$SSIM} 
    & \multicolumn{1}{c|}{$\downarrow$LPIPS} 
    & \multicolumn{1}{c}{$\uparrow$PSNR} 
    & \multicolumn{1}{c}{$\uparrow$SSIM} 
    & \multicolumn{1}{c|}{$\downarrow$LPIPS} 
    & \multicolumn{1}{c}{$\uparrow$PSNR} 
    & \multicolumn{1}{c}{$\uparrow$SSIM} 
    & \multicolumn{1}{c||}{$\downarrow$LPIPS}  
    & \multicolumn{1}{c}{$\downarrow$Time (h)} \\
    \midrule
    Plenoxels~\cite{fridovich2022plenoxels}
    &17.75  & 0.419 & 0.670
    &20.48  & 0.462 & 0.658
    &18.27  & 0.517 & 0.579
    &18.93  & 0.638 & 0.528
    &20.40  & 0.456 & 0.780
    &15.76  & 0.520 & 0.665 & {01:30} \\
    TensoRF (VM-192)~\cite{chen2022tensorf}\xspace 
    &18.19 &0.416 &0.697 
    &20.77 &0.452 &0.675 
    &18.32 &0.498 &0.607  
    &18.39 &0.594 &0.585 
    &17.22 &0.418 &0.842 
    &14.56 &0.512 &0.679 & 01:31 \\
    Instant-NGP~\cite{muller2022instant}\xspace 
    &18.68 &0.439 &0.614 
    &21.18 &0.497 &0.579
    &18.91 &0.554 &0.543 
    &19.00 &0.644 &0.528
    &16.31 &0.527 &0.729
    &14.91 &0.509 &0.732 &\textbf{01:22} \\
    Mega-NeRF*~\cite{turki2022mega}  \xspace 
    & {20.93} & {0.547}   & {0.504}
    & {24.06} & {0.553}   & {0.516}
    & {22.08} & {0.628}   & {0.489}
    & \textbf{25.60} & {0.770}   & {0.390}
    & {23.42} & {0.537}   & {0.618}
    & {17.66} & \textbf{0.536}   & \textbf{0.616} & 20:10 \\
    Ours\xspace 
    & \textbf{20.99} & \textbf{0.565}   & \textbf{0.490} 
    & \textbf{24.08} & \textbf{0.563}   & \textbf{0.497} 
    & \textbf{22.41} & \textbf{0.659}   & \textbf{0.451} 
    & {25.56} & \textbf{0.783}   & \textbf{0.373}  
    & \textbf{23.46} & \textbf{0.544}   & \textbf{0.611} 
    & \textbf{17.67} & {0.521}   & {0.623} & 01:35   \\

    \bottomrule
\end{tabular}
}

\end{table*}

\paragraph{Independent Background Modeling.} 
Existing methods show that representing the foreground and background regions independently leads to better results~\cite{Zhang20arxiv_nerf++,turki2022mega}.
To increase the model capacity, we utilize two separate models for the foreground and background regions.
For the foreground region, we use the proposed hybrid feature representation for feature encoding.
For the background region, we only use a \hashgrid with a hash table size of $T_{bg}=2^{19}$.
Our design can improve the performance without an obvious increase of time costs. 
Then, points in different regions will be processed by the corresponding models.

\paragraph{Loss Function.} 
Given multiple 3D points sampled from a view ray $\ray$, volume rendering will be utilized to produce the pixel color.
We adopt the mean square error (MSE) between the rendered color $\hat{\raycolor}(\ray)$ and the ground-truth color $\raycolor(\ray)$ as the loss function:
\begin{align}
    \label{eq:loss function}
     Loss = \sum_{\ray \in \mathcal{R}}  {\| \hat{\raycolor}(\ray) - \raycolor(\ray)\|}_{2}^{2},
\end{align}
where $\mathcal{R}$ is the sampled ray set.

\section{Experiments}%

\paragraph{Datasets.} Following Mega-NeRF~\cite{turki2022mega}, we evaluate our \MethodName on three public large-scale datasets, namely the Mill19~\cite{turki2022mega}, Quad6k~\cite{crandall2011discrete}, and UrbanScene3D~\cite{UrbanScene3D} datasets, with the camera poses refined by PixSFM~\cite{lindenberger2021pixel}.
Mill19 dataset is proposed by Mega-NeRF~\cite{turki2022mega}, which includes Mill19-Building and Mill19-Rubble, covering about $250 \times 400 \ m^2$ large areas. Quad6k dataset is a large-scale Structure-from-Motion (SfM) dataset containing about 5100 images of Cornell University  Arts Quad.  
Following~\cite{turki2022mega}, we adopt three urban scenes from UrbanScene3D dataset, which provide high-resolution images from drones.

\paragraph{Implementation Details.} 
We implemented our method with PyTorch~\cite{paszke2017pytorch} and used the Adam optimizer~\cite{kingma2015adam} with default parameters. 
Our model was trained with $100$K iteration using a batch size of $5\times 1024$ rays. The learning rate was set to 0.001. 
Following the practice in~\cite{muller2022instant}, the hyper-parameter of hash table size is set to $L=16, T=2^{19}, \featnum=2$. The lowest and highest resolutions are set to $16$ and $2048$, respectively.
The runtime was measured on an NVIDIA GeForce RTX 3090 GPU.

\paragraph{Evaluation Metrics.}
We evaluate the existing methods and our method in terms of PSNR, \zyqm{SSIM}~\cite{wang2004image}, and the VGG implementation of LPIPS~\cite{zhang2018unreasonable}.

\label{sec:Experiments}
\begin{table*}[t] \centering
    \newcommand{\Frst}[1]{\textcolor{red}{\textbf{#1}}}
    \newcommand{\Scnd}[1]{\textcolor{blue}{\textbf{#1}}}
    \captionof{table}{Comparison between our hybrid-representation and existing feature representations. We compare with three baselines that replace the hybrid-representation in our method with the dense-grid~\cite{sun2022direct,fridovich2022plenoxels}, the pure \hashgrid~\cite{muller2022instant}, and the TensoRF (VM-192)~\cite{chen2022tensorf}.}
    \label{tab:quantitative_baseline}
    
\makebox[\textwidth]{} 
\resizebox{\textwidth}{!}{
\begin{tabular}{l||*{3}{c}|*{3}{c}|*{3}{c}|*{3}{c}|*{3}{c}|*{3}{c}||c}
    \toprule
    & \multicolumn{3}{c|}{Mill 19 - Building} 
    & \multicolumn{3}{c|}{Mill 19 - Rubble}
    & \multicolumn{3}{c|}{UrbanScene3D - Residence}
    & \multicolumn{3}{c|}{UrbanScene3D - Sci-Art}
    & \multicolumn{3}{c|}{UrbanScene3D - Campus}
    & \multicolumn{3}{c||}{Quad 6k}
    & \multicolumn{1}{c}{Timing}\\
    Method
    & \multicolumn{1}{c}{$\uparrow$PSNR} 
    & \multicolumn{1}{c}{$\uparrow$SSIM} 
    & \multicolumn{1}{c|}{$\downarrow$LPIPS} 
    & \multicolumn{1}{c}{$\uparrow$PSNR} 
    & \multicolumn{1}{c}{$\uparrow$SSIM} 
    & \multicolumn{1}{c|}{$\downarrow$LPIPS} 
    & \multicolumn{1}{c}{$\uparrow$PSNR} 
    & \multicolumn{1}{c}{$\uparrow$SSIM} 
    & \multicolumn{1}{c|}{$\downarrow$LPIPS} 
    & \multicolumn{1}{c}{$\uparrow$PSNR} 
    & \multicolumn{1}{c}{$\uparrow$SSIM} 
    & \multicolumn{1}{c|}{$\downarrow$LPIPS} 
    & \multicolumn{1}{c}{$\uparrow$PSNR} 
    & \multicolumn{1}{c}{$\uparrow$SSIM} 
    & \multicolumn{1}{c|}{$\downarrow$LPIPS} 
    & \multicolumn{1}{c}{$\uparrow$PSNR} 
    & \multicolumn{1}{c}{$\uparrow$SSIM} 
    & \multicolumn{1}{c||}{$\downarrow$LPIPS}  
    & \multicolumn{1}{c}{$\downarrow$Time (h)} \\
    \midrule
    Ours w/ TensoRF~\cite{chen2022tensorf} \xspace 
    &19.44  & 0.457 & 0.615
    &22.82  & 0.485 & 0.621 
    & 21.17 & 0.573 & 0.552 
    & 22.59 & 0.656 & 0.535 
    & 22.05 & 0.487 & 0.720
    & 15.15 & 0.517 & 0.666 & 03:34 \\
    Ours w/ \densegrid~\cite{sun2022direct,fridovich2022plenoxels}\xspace 
    & 18.95 & 0.437 & 0.636 
    & 22.46 & 0.464 & 0.649 
    & 20.43 & 0.541 & 0.571 
    & 21.24 & 0.629 & 0.560 
    & 21.61 & 0.475 & 0.730 
    & 14.61 & 0.508 & 0.687 & 01:31 \\
    Ours w/ \hashgrid~\cite{muller2022instant}\xspace 
    & 20.58 & 0.539 & 0.514    
    & 23.77 & 0.546 & 0.516 
    & 22.05 & 0.634 & 0.478  
    & 25.06 & 0.770 & 0.391   
    & 22.96 & 0.528 & 0.638 
    & 17.31 & 0.516 & 0.636 & \textbf{01:21} \\
    Ours w/ \hashgrid (01:30)\xspace 
    & 20.66 & 0.545         & 0.509 
    & 23.77 & 0.548         & 0.515
    & 22.06 & 0.636  & 0.477 
    & 25.15 & 0.772  & 0.388 
    & 23.07 & 0.530         & 0.633 
    & 17.34 & 0.516         & 0.631 & 01:34 \\
    Ours\xspace 
    & \textbf{20.99} & \textbf{0.565}   & \textbf{0.490} 
    & \textbf{24.08} & \textbf{0.563}   & \textbf{0.497} 
    & \textbf{22.41} & \textbf{0.659}   & \textbf{0.451} 
    & \textbf{25.56} & \textbf{0.783}   & \textbf{0.373}  
    & \textbf{23.46} & \textbf{0.544}   & \textbf{0.611} 
    & \textbf{17.67} & \textbf{0.521}   & \textbf{0.623} 
    & 01:35   \\
    \bottomrule
\end{tabular}
}

    \makebox[0.160\textwidth]{\footnotesize (a) Ours w/ \densegrid}
    \makebox[0.160\textwidth]{\footnotesize (b) Ours w/ TensoRF}
    \makebox[0.160\textwidth]{\footnotesize (c) Ours w/ \hashgrid}
    \makebox[0.160\textwidth]{\footnotesize (d) \MethodName (ours)}
    \makebox[0.160\textwidth]{\footnotesize (e) Mega-NeRF}
    \makebox[0.160\textwidth]{\footnotesize (f) Ground Truth}
    \\
    \includegraphics[width=0.160\textwidth]{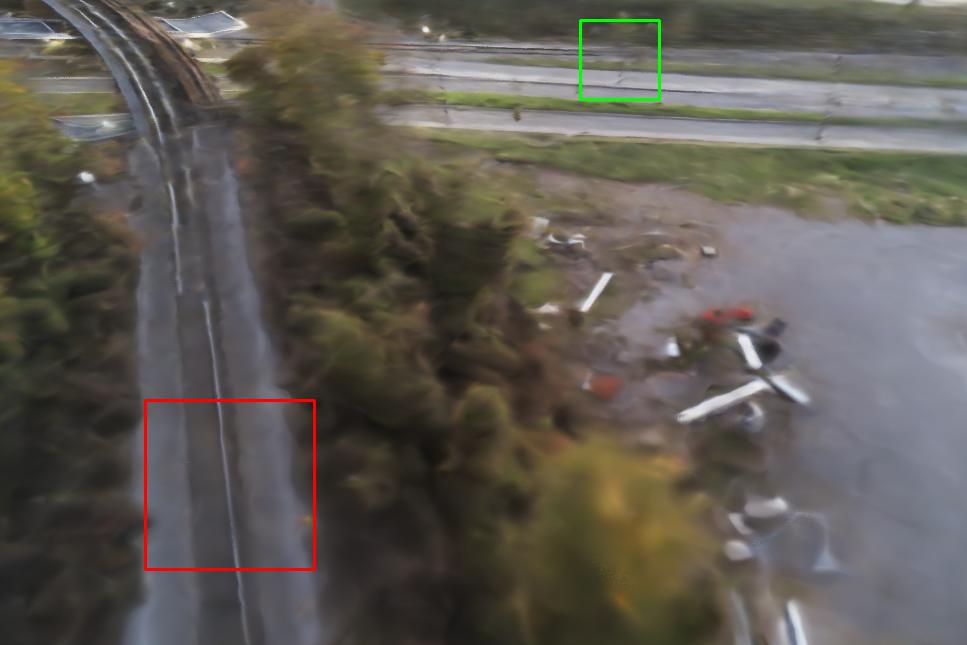}
    \includegraphics[width=0.160\textwidth]{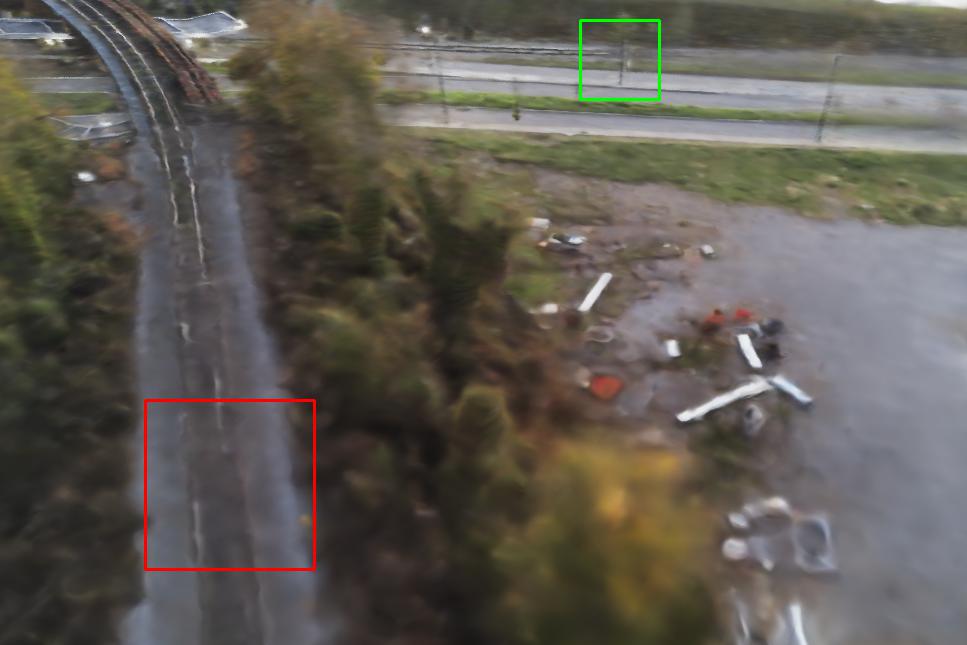}
    \includegraphics[width=0.160\textwidth]{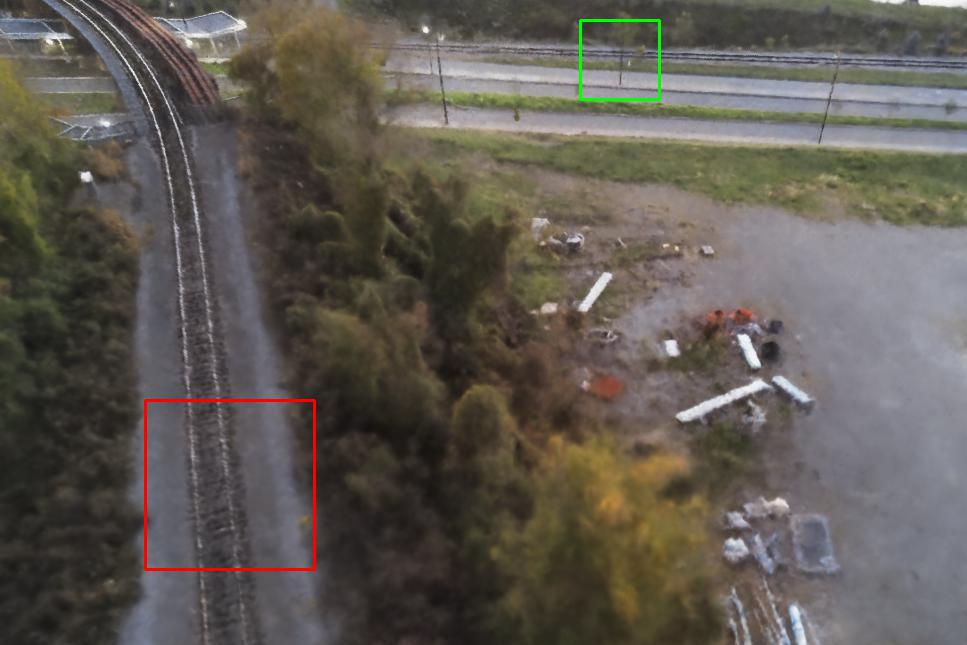}
    \includegraphics[width=0.160\textwidth]{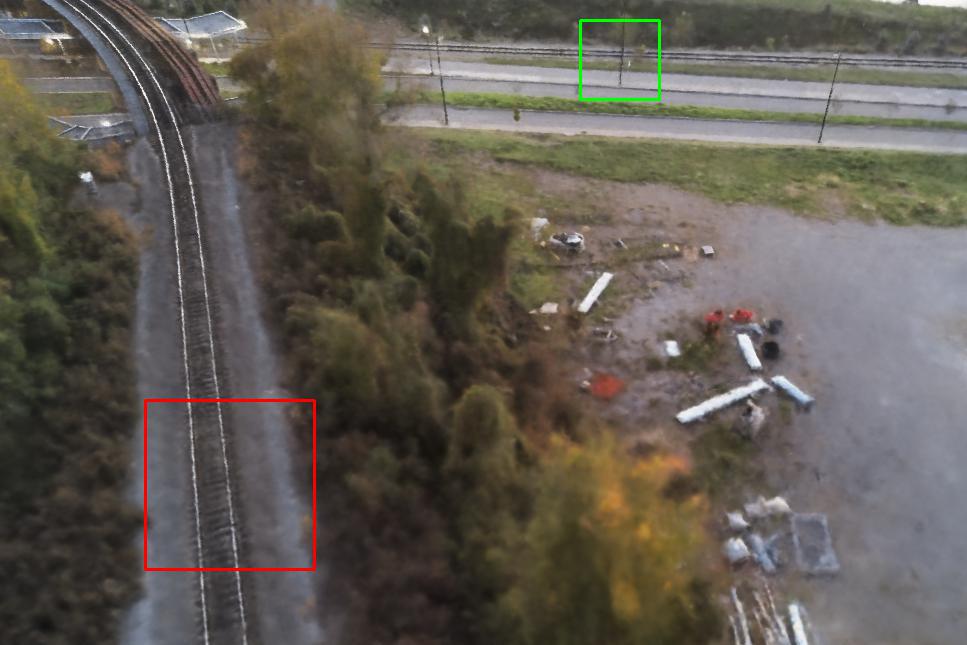}
    \includegraphics[width=0.160\textwidth]{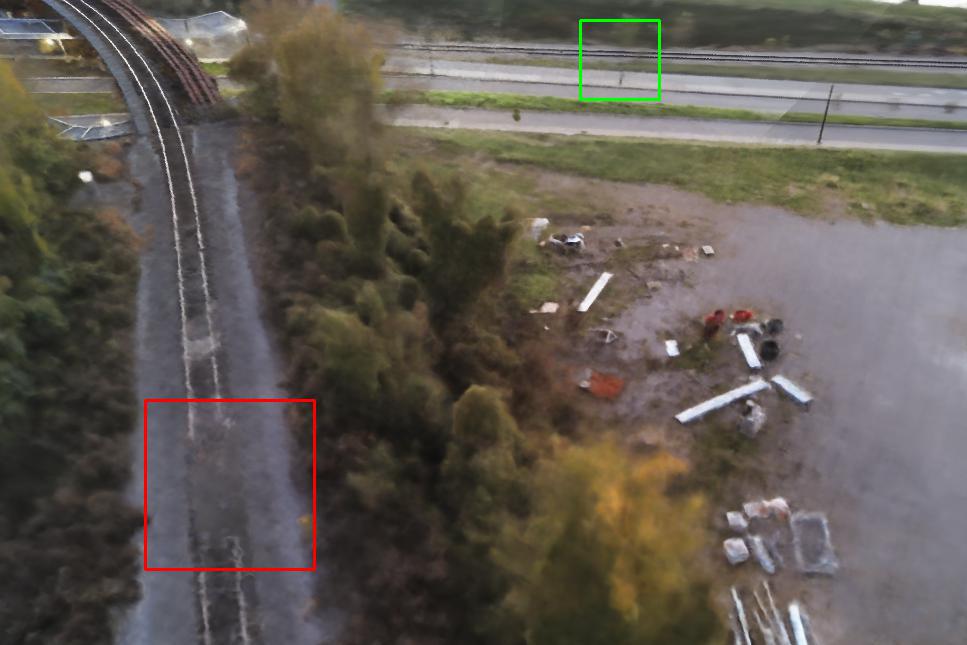}
    \includegraphics[width=0.160\textwidth]{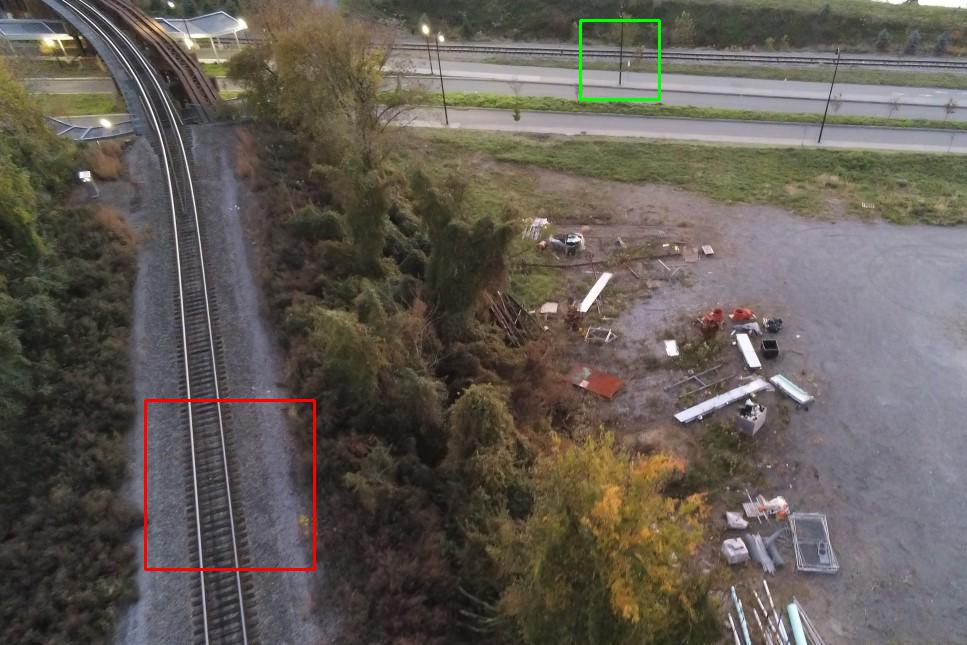}
    \\
    \includegraphics[width=0.0775\textwidth]{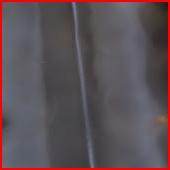}
    \includegraphics[width=0.0775\textwidth]{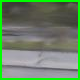}
    \includegraphics[width=0.0775\textwidth]{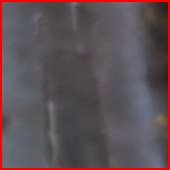}
    \includegraphics[width=0.0775\textwidth]{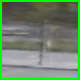}
    \includegraphics[width=0.0775\textwidth]{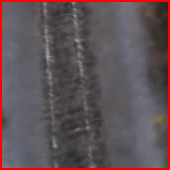}
    \includegraphics[width=0.0775\textwidth]{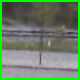}
    \includegraphics[width=0.0775\textwidth]{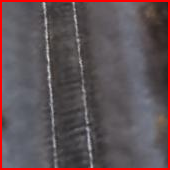}
    \includegraphics[width=0.0775\textwidth]{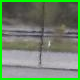}
    \includegraphics[width=0.0775\textwidth]{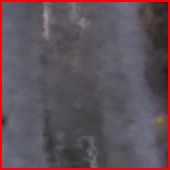}
    \includegraphics[width=0.0775\textwidth]{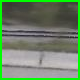}
    \includegraphics[width=0.0775\textwidth]{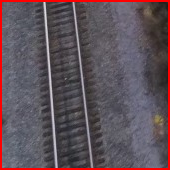}
    \includegraphics[width=0.0775\textwidth]{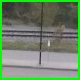}
    \\ 
    \includegraphics[width=0.160\textwidth]{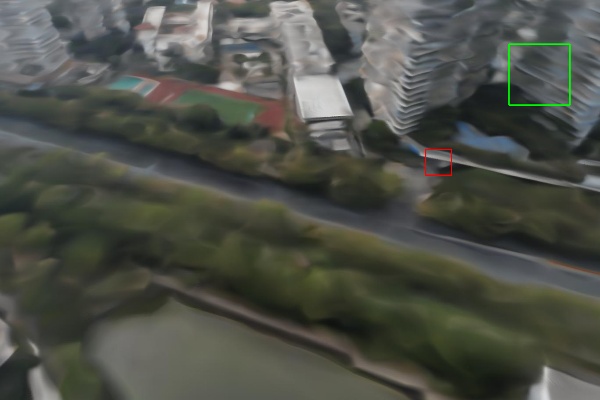}
    \includegraphics[width=0.160\textwidth]{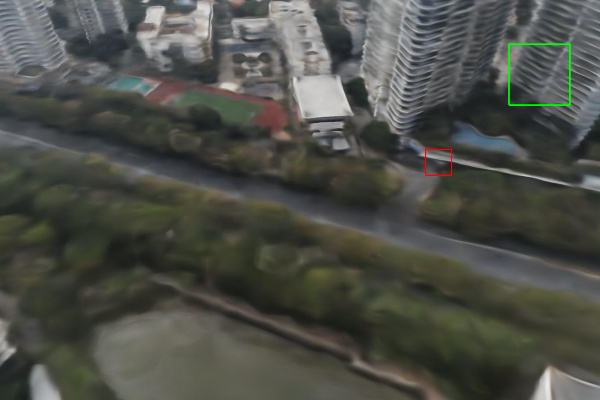}
    \includegraphics[width=0.160\textwidth]{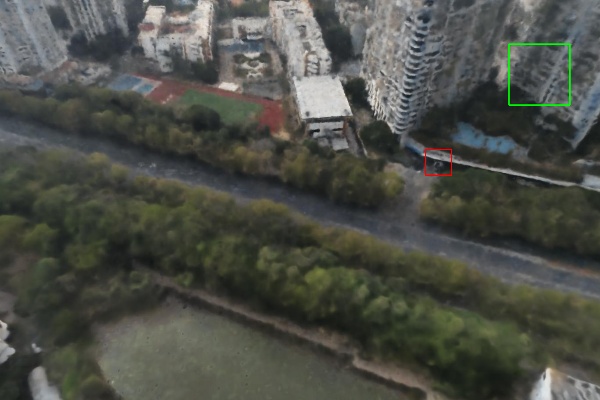}
    \includegraphics[width=0.160\textwidth]{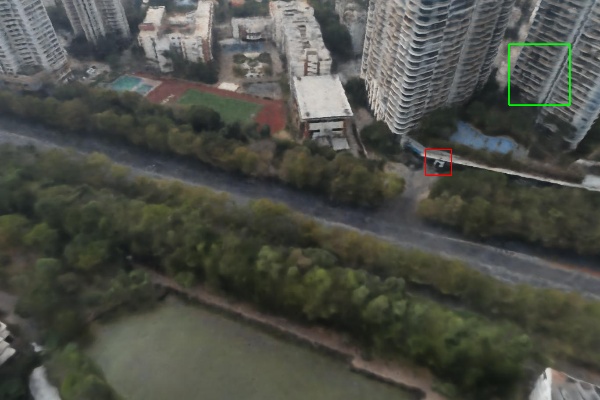}
    \includegraphics[width=0.160\textwidth]{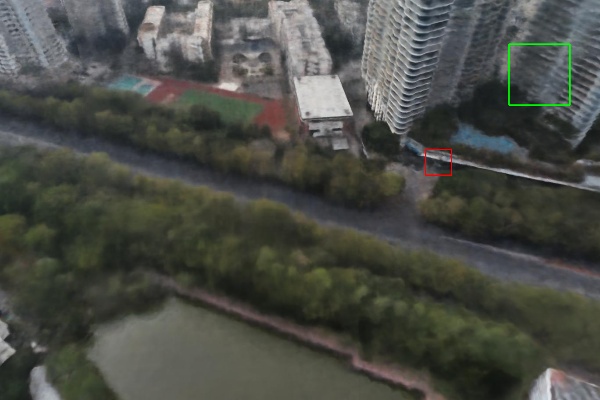}
    \includegraphics[width=0.160\textwidth]{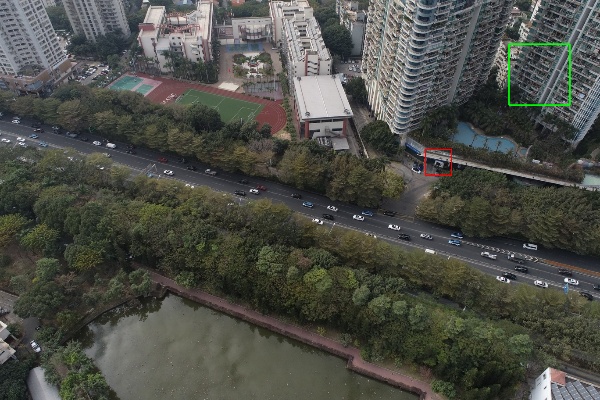}
    \\
    \includegraphics[width=0.0775\textwidth]{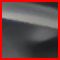}
    \includegraphics[width=0.0775\textwidth]{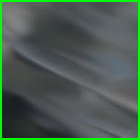}
    \includegraphics[width=0.0775\textwidth]{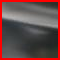}
    \includegraphics[width=0.0775\textwidth]{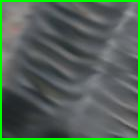}
    \includegraphics[width=0.0775\textwidth]{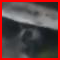}
    \includegraphics[width=0.0775\textwidth]{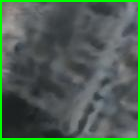}
    \includegraphics[width=0.0775\textwidth]{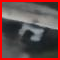}
    \includegraphics[width=0.0775\textwidth]{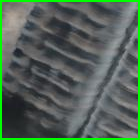}
    \includegraphics[width=0.0775\textwidth]{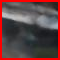}
    \includegraphics[width=0.0775\textwidth]{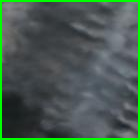}
    \includegraphics[width=0.0775\textwidth]{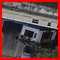}
    \includegraphics[width=0.0775\textwidth]{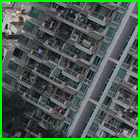}

    \captionof{figure}{\textbf{Qualitative results} on the Mill19-Rubble and UrbanScene3D-Campus dataset.} 
    \label{fig:Qualitative}
\end{table*}

\subsection{Evaluation}%
\label{sub:Evaluation}

\paragraph{Comparison with Existing Methods.}
We first compare our method with existing methods to verify its effectiveness. 
\gym{Specifically, we compare with the state-of-the-art Mega-NeRF~\cite{turki2022mega} and fast NeRF methods (Plenoxels~\cite{fridovich2022plenoxels}, TensoRF~\cite{chen2022tensorf}, and Instant-NGP~\cite{muller2022instant}).
The training time are measured by training on the same amount of data.}

\gym{We can see from~\Tref{tab:quantitative} that our method achieves the best results on five of the six scenes.}
Compared with Mega-NeRF which requires about one day's training with 8 GPUs, our method achieves a fast \zyqm{convergence} speed of 1.5 hours using a single GPU, while maintaining comparable or even better results.  
The reason is that our method only use a single 5-layer network with 64 channels and an efficient hybrid feature representation for feature interpolation, while Mega-NeRF uses large 9-layer MLPs with 256 channels for predictions. 

\gym{Despite enjoying fast training convergence, existing fast NeRF methods~\cite{fridovich2022plenoxels,chen2022tensorf,muller2022instant} achieve poor rendering quality. It is because these methods are not designed for large-scale unbounded scenes with illumination changes and suffer from low-resolution representation.}

\paragraph{Comparison with Other Feature Representations.} 
\gym{To further demonstrate the effectiveness of our hybrid feature representation,
we design three baseline methods.} 
Specifically, we replace the proposed hybrid feature representation of our method with the dense-grid (resolution is $160^3$)~\cite{sun2022direct,fridovich2022plenoxels} and pure \hashgrid~\cite{muller2022instant} representation. 
In addition, we adopt the VM-192 variant of TensoRF with a resolution of $300^3$ voxels~\cite{chen2022tensorf}. 
To have a fair comparison, other parts of the methods are kept the same.
We also apply the feature space scaling according to the camera altitude measurement for the baseline methods for fair comparisons.
Quantitative results in \Tref{tab:quantitative_baseline} show that the results of methods with dense-grid and TensoRF representation are worse, indicating that they are not suitable for large-scale scenes as their feature resolutions are low.  
Compared with these two methods, our method achieves much more accurate rendering results, with comparable or smaller parameters (\ie, ours is 33.0M, \emph{Ours w/ TensoRF} is 34.8M, and \emph{Ours w/ \densegrid} is 98.4M).

Our method also \gym{consistently} outperforms \emph{Ours w/ \hashgrid} on all six datasets.
We notice that the \hashgrid method finishes training with less time, we therefore design experiments to train the \hashgrid method with the same time as ours for a fair comparison, namely \emph{Ours w/ \hashgrid (01:30)} in \Tref{tab:quantitative_baseline}. The results show that the performance of \hashgrid method improves with a longer training time, but its results are still far behind our method, further demonstrating the effectiveness of our method.

\Fref{fig:Qualitative} shows the qualitative comparison. We can see that the baselines based on low-resolution features, \eg \emph{Ours w/ TensoRF} and \emph{\densegrid}, produce blurry results, while \zyqm{\emph{Ours w/ \hashgrid}} produces better results with high-resolution hash-grid. 
However, at the presence of collision, the rendering details are still blurry. In contrast, we design the multi-scale high-resolution dense plane and utilize the hybrid-representation to enhance the \hashgrid feature, leading to better results (\eg the railway track in the Mill19-Rubble dataset).

\subsection{Ablation study}%
\label{sub:Ablation study}
In this section, we conduct ablation studies to evaluate the proposed \MethodName on two representative scenes, Mill19-Rubble~\cite{turki2022mega} and UrbanScene3D-Residence~\cite{UrbanScene3D}.

\begin{figure}[t] \centering
    \includegraphics[width=0.235\textwidth]{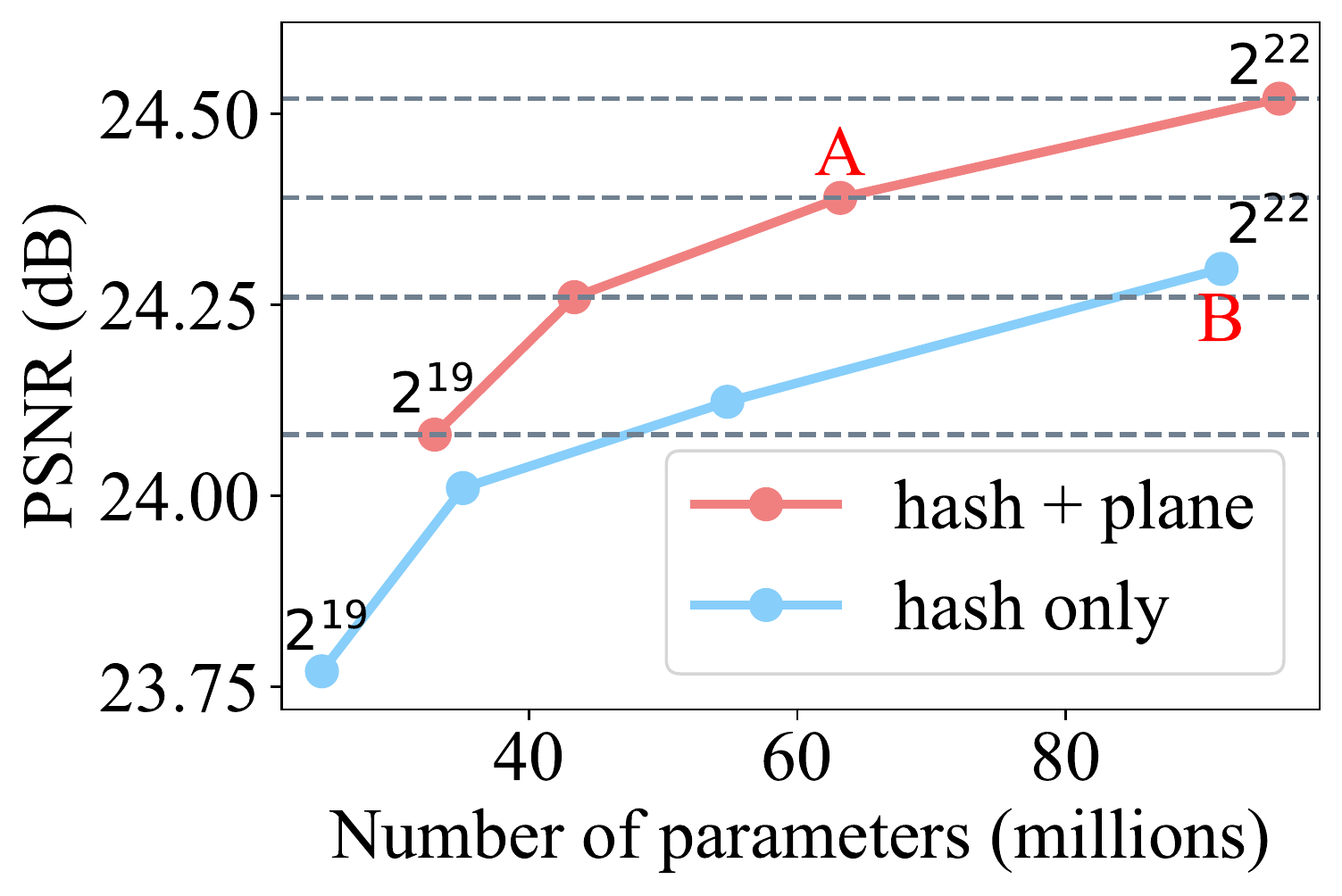}
    \includegraphics[width=0.235\textwidth]{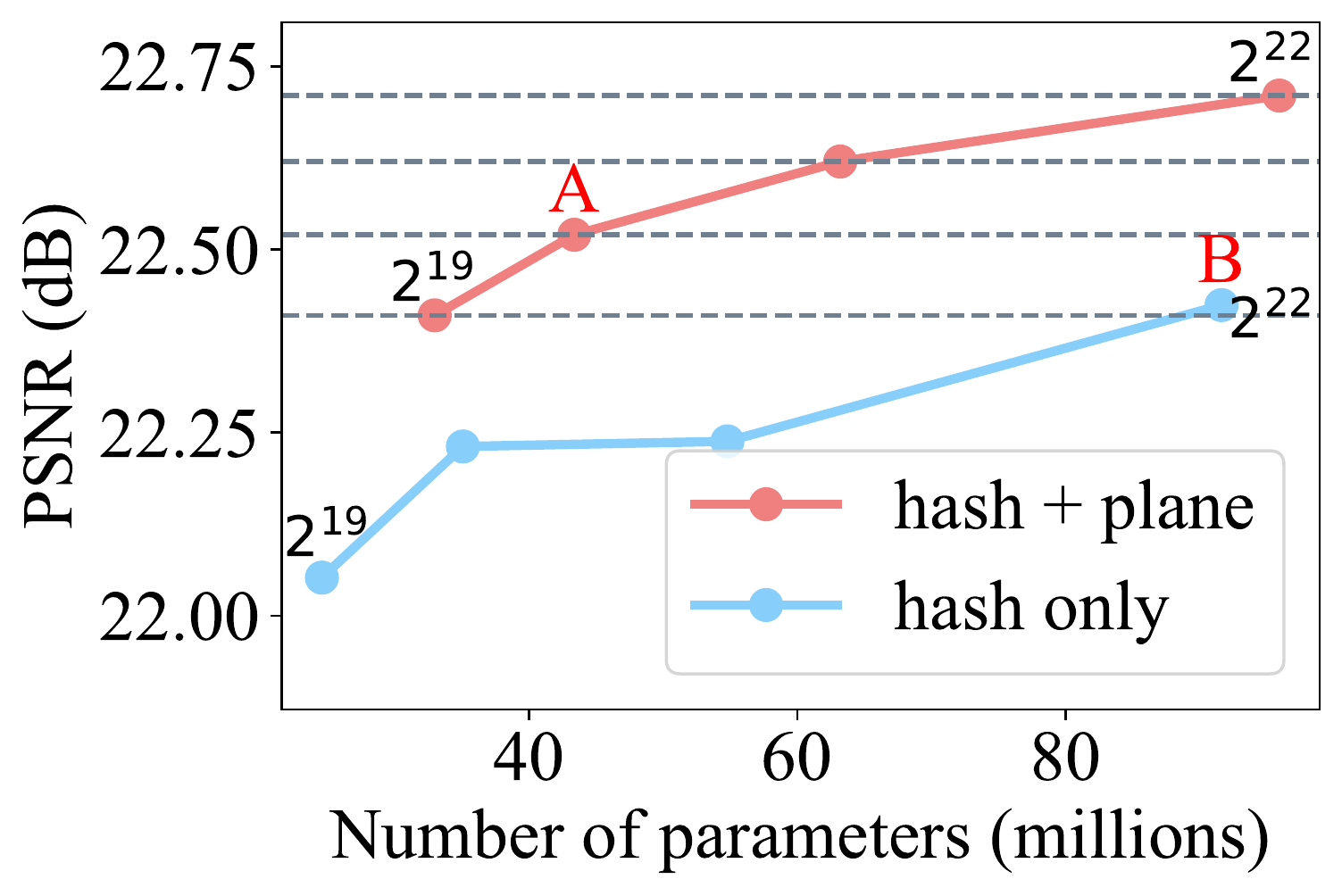}
    \\
    \vspace{-6pt}
    \makebox[0.235\textwidth]{\scriptsize (a) Rubble}
    \makebox[0.235\textwidth]{\scriptsize (b) Residence}
    \caption{Quantitative results of integrating the plane feature with hash-grids that have a hash table size $T$ ranged from $2^{19}$ to $2^{22}$.}  
    \label{fig:hash plane}
\end{figure}

\paragraph{Effectiveness of the Plane Feature.} 
To verify the effectiveness of the proposed hybrid feature representation, we integrate our plane features with hash-grid with different hash table sizes (\ie, from $2^{19}$ to $2^{22}$) and compare them with methods using only hash-grid features, and the results are summarized in \fref{fig:hash plane}.
We can see that as the hash table size improves, the results of hash-grid only method increase but with a significant increase in model size. 
It can be clearly seen that the performance of models with a hybrid of \hashgrid feature (with different hash table sizes) and plane features exceed the corresponding pure \hashgrid results. 
Notably, the hybrid representation can achieve better results using much fewer number of parameters than a pure \hashgrid with higher table size (\eg, comparing the point $A$ and $B$ in \fref{fig:hash plane}).

\begin{figure}[tb] \centering
    \input{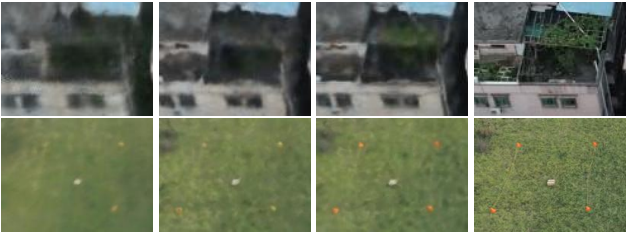}
    \caption{Effectiveness of the plane feature.} 
    \vspace{-10pt}
    \label{fig:planeonly}
\end{figure}

To better verify the effectiveness of our plane feature, we remove the \hashgrid and only use plane features for feature encoding. \zyqm{Experiment with ID 0 in \Tref{tab:ablation component} shows} that the pure plane features can also achieve respectable results \gym{(see~\fref{fig:planeonly} for visual results)}. 
\zyqm{Furthermore, we replace the plane feature in our hybrid representation with the dense-grid ($90^3$) and TensoRF (VM-192, $210^3$) that has comparable parameters as our method (\ie, 33.0M).
Experiments with IDs 1-4 in \Tref{tab:ablation component}
shows that integrating \hashgrid with the low-resolution dense-grid or TensoRF cannot consistently improve the results and might lead to worse results (see the average PSNR), indicating that the high-resolution plane features help to resolve the collision problem and the improvement comes from our design rather than an increase of parameters.}
Both the quantitative and qualitative results reiterate that the plane features provide strong information for the scene geometry and appearance and is complementary to the \hashgrid features.

\begin{table}[tb]\centering
    \caption{Effectiveness of the model design.}
    \label{tab:ablation component}
    \resizebox{0.48\textwidth}{!}{
    \footnotesize
    \begin{tabular}{*{1}{c}*{1}{l}|*{1}{c}*{2}{c}*{2}{c}*{2}{c}}
        \toprule
        & & & \multicolumn{2}{c}{Rubble} & \multicolumn{2}{c}{Residence} & \multicolumn{2}{c}{Average}\\
       ID & Model & \# Param & PSNR & SSIM & PSNR & SSIM  & PSNR & SSIM \\
        \midrule
        0 
        & Plane only   & {16.8M} & 22.75 & 0.493 & 20.61 & 0.554  & 21.68 & 0.524\\
        \rowcolor{gray!20}1 
        & Hash-grid ($2^{19}$) & 24.6M  & 23.77 & 0.546 & 22.05 & 0.634  & 22.91 & 0.590 \\
        2
        &Hash-grid + Dense & 33.3M & 23.82  &0.554  &21.86  &0.639  &22.84$\downarrow$ & 0.597 \\
        3
        &Hash-grid + TensoRF & 33.1M & 23.77  &0.547  &22.02  &0.634    &22.90$\downarrow$ & 0.591 \\
        4 
        & Hash-grid + Plane (Ours) & 33.0M & \textbf{24.08} & \textbf{0.565} & \textbf{22.41} & \textbf{0.659}  & \textbf{23.25}$\uparrow$ & \textbf{0.612}\\
        \bottomrule
    \end{tabular}
}

\end{table}

\paragraph{Effects of Plane Feature Dimension.}
In \Tref{tab:feature dim and multi-scale}, we first conduct experiments with IDs 0-1 to investigate the influence of feature dimension in a single-resolution plane representation. 
We can see that the result of a single-resolution plane with small feature dimension (\eg, $2$) is worse, and using a high feature dimension (\eg, $8$) can achieve good PSNR results, but the increase in the number of parameters is also significant. 
Experiments with ID 2-3 in \Tref{tab:feature dim and multi-scale} show that multi-resolution planes with small resolution can also achieve good results while maintaining a small parameter number.
These results justify the design of our multi-resolution plane features, which strikes a good balance between the performance and parameters.

\paragraph{Effects of Plane Feature Resolution.}
To investigate the influence of the resolution of plane features, we compare results of plane features with different highest resolutions (the base resolution is $128$).  
\Tref{tab:hash plane} shows that there is a tendency that the larger the highest resolution is, the better performance achieves, which is a trade-off between the increase of parameters and performance. This also demonstrates the effectiveness of plane features.

\begin{table}[t]\centering
    \caption{Effects of the feature dimension per resolution and the multi-resolution design in the plane feature representation.}
    \label{tab:feature dim and multi-scale}
    \resizebox{0.48\textwidth}{!}{
    \begin{tabular}{*{2}{c}|*{2}{c}*{10}{c}}
        \toprule
        &  &  &  & \multicolumn{2}{c}{Rubble} & \multicolumn{2}{c}{Residence} \\
        IDs & Resolutions / Dim. & Total Dim. & \# Param & PSNR & SSIM & PSNR & SSIM\\
        \midrule
        0 & \{1024\} / 2 & 2 
        & {30.9M} & 23.92 & 0.556 & 22.32 & 0.652\\
        
        1 & \{1024\} / 8 & 8 
        & 49.8M & \textbf{24.14} & \textbf{0.570} & \textbf{22.60} & \textbf{0.670}\\
        
        2 & \{128,1024\} / 4 & 8 
        & 37.4M & 24.10 & 0.566 & 22.41 & 0.661 \\
        
        \rowcolor{gray!20}3 & \{128,256,512,1024\} / 2 & 8 
        & 33.0M & 24.08 & 0.565 & 22.41 & 0.659 \\
        \bottomrule
    \end{tabular}
    }

    \vspace{1em}
    \vspace{0.07em}
    \caption{Effects of the highest resolution in plane features.}
    \label{tab:hash plane}

\resizebox{0.48\textwidth}{!}{
\footnotesize
\begin{tabular}{*{1}{c}|*{1}{c}*{10}{c}}
    \toprule
   & & \multicolumn{2}{c}{Rubble} & \multicolumn{2}{c}{Residence} \\
   Model & \# Param & PSNR & SSIM & PSNR & SSIM \\
    
    \midrule
    \Hashgrid $2^{19}$ + plane 256 
    & {25.5M} & 23.97 & 0.557 & 22.20 & 0.643 \\
    \Hashgrid $2^{19}$ + plane 512 
    & 27.1M & 24.04 & 0.559 & 22.16 & 0.648 \\
    \rowcolor{gray!20}\Hashgrid $2^{19}$ + plane 1024 
    & 33.0M & 24.08 & 0.565 & 22.41 & 0.659\\
    \Hashgrid $2^{19}$ + plane 2048
    & 54.5M & \textbf{24.32} & \textbf{0.589} & \textbf{22.43} & \textbf{0.673} \\
    \bottomrule
\end{tabular}}

    \vspace{1em}
    \vspace{0.07em}
    \caption{Results of Mega-NeRF variants trained with 8 GPUs.}
        \resizebox{0.48\textwidth}{!}{
    \footnotesize
    \begin{tabular}{*{1}{l}|*{2}{c}*{2}{c}*{2}{c}*{2}{c}}
        \toprule
        & \multicolumn{2}{c}{Building} & \multicolumn{2}{c}{Rubble} & \multicolumn{2}{c}{Residence}  & \multicolumn{2}{c}{Campus} \\
       Model  & PSNR & SSIM & PSNR & SSIM & PSNR & SSIM & PSNR & SSIM\\
        \midrule
        Mega-NeRF (1 day) 
        & 20.93 & 0.547
        & 24.06 & 0.553
        & 22.08 & 0.628 
        & 23.42 & 0.537\\
        \zyqm{Mega-NeRF (1.5 h)}
        & 19.10 & 0.436
        & 22.29 & 0.462
        & 20.08 & 0.516 
        & 21.67 & 0.476\\
        \midrule
        Mega-NeRF w/ \zyqm{INGP (1.5 h)} 
        & 21.30 & 0.601
        & 24.47 & 0.607
        & 22.36 & 0.671
        & 23.59 & 0.565\\
        Mega-NeRF w/ \zyqm{GP-NeRF (1.5 h)}
        & \textbf{21.71} & \textbf{0.619}
        & \textbf{24.80} & \textbf{0.623}
        & \textbf{22.81} & \textbf{0.698}
        & \textbf{23.94} & \textbf{0.580}\\
        \bottomrule
    \end{tabular}
    }

    \label{tab:mega_variants}
    \vspace{-0.5em}
\end{table}

\subsection{GP-NeRF as the Local Radiance Field}
\gym{As a general NeRF variant, our GP-NeRF can also be adopted as the local radiance field in Mega-NeRF's partition training framework~\cite{turki2022mega} to further improve the performance. 
We trained two Mega-NeRF variants
with GP-NeRF and Instance-NGP~\cite{muller2022instant} (for purpose of comparison) as the local NeRF.
\Tref{tab:mega_variants} shows that 
 adopting our GP-NeRF as the local radiance field achieves the best results in $1.5$ hours, clearly outperforms Mega-NeRF and the Instant-NGP variant, verifying the design of our method.
}

\section{Conclusion}%
\label{sec:Conclusion}
In this paper, we have presented a hybrid feature representation for the neural radiance fields to enable efficient large-scale scene modeling.
Our hybrid representation enhances the \hashgrid feature with orthogonally placed high-resolution and dense plane features, especially for surface regions with hash collisions.
Compared with directly scaling up the hash table size in the \hashgrid, our representation can achieve higher accuracy with comparable runtime while using much fewer parameters.
Based on our hybrid representation, we propose a new variant of NeRF, called \MethodName, to represent large-scale \zyqm{unbounded} scenes. 
Compared with the existing method~\cite{turki2022mega} that requires one day's training on 8 GPUs, our method achieves comparable or even better results with $1.5$ hours on a single GPU, which is especially useful for scenarios with limited computing resources.

\vspace{0.5em}
\paragraph{Limitations.} Despite the significant acceleration of optimization achieved for large-scale scene representation, 
our method still cannot achieve real-time scene reconstruction.
Moreover, our method does not explicitly model dynamic objects and thus cannot represent a dynamic scene. 
We would like to address these two problems in the future.

\clearpage
{\small
\bibliographystyle{ieee_fullname}
\bibliography{ref}
}

\end{document}